\documentclass[10pt,twocolumn,letterpaper]{article}

\usepackage{cvpr}
\usepackage{times}
\usepackage{epsfig}
\usepackage{graphicx}
\usepackage{amsmath}
\usepackage{amssymb}
\usepackage{multirow}
\usepackage{booktabs}
\usepackage{makecell}
\usepackage[colorlinks,linkcolor=blue,anchorcolor=blue,citecolor=green,]{hyperref}

\cvprfinalcopy 

\def\cvprPaperID{****} 

\begin{document}

\title{Temporal Fusion Network for Temporal Action Localization: \\
Submission to ActivityNet Challenge 2020 (Task E)}

\author{Zhiwu Qing$^1$ \quad  \quad Xiang Wang$^1$ \quad Yongpeng Sang$^2$ \quad Changxin Gao$^1$ \\
\quad Shiwei Zhang$^{3*}$ \quad Nong Sang$^1$\thanks{
Corresponding authors
}
\\
$^1$School of Artificial Intelligence and Automation, Huazhong University of Science and Technology\\
$^2$School of Cyber Science and Engineering, Huazhong University of Science and Technology\\
$^3$DAMO Academy, Alibaba Group\\
{\tt\small \{qzw, u201613707, ypsang, cgao, nsang\}@hust.edu.cn}\\
{\tt\small zhangjin.zsw@alibaba-inc.com}
}

\maketitle

\begin{abstract}
   This technical report analyzes a temporal action localization method we used in the HACS competition which is hosted in Activitynet Challenge 2020.
   The goal of our task is to locate the start time and end time of the action in the untrimmed video, and predict action category.
   Firstly, we utilize the video-level feature information to train multiple video-level action classification models. In this way, we can get the category of action in the video.
   Secondly, we focus on generating high quality temporal proposals.
   For this purpose, we apply BMN to generate a large number of proposals to obtain high recall rates. 
   We then refine these proposals by employing a cascade structure network called Refine Network, which can predict position offset and new IOU under the supervision of ground truth.
   To make the proposals more accurate, we use bidirectional LSTM, Nonlocal and Transformer to capture temporal relationships between local features of each proposal and global features of the video data.
   Finally, by fusing the results of multiple models, our method obtains $40.55\%$ on the validation set and $40.53\%$ on the test set in terms of mAP, and achieves \textbf{Rank $1$} in this challenge.
\end{abstract}


\section{Our Approach}

Inspired by current state-of-the-art method~\cite{lin2019bmn}, we decouple the task of temporal action localization into two subtasks, \emph{i.e.}, video classification and  proposal generation.
First of all, we use Slowfast-101~\cite{feichtenhofer2019slowfast} as a backbone to train a video-level action classification model.
We then use the trained backbone to extract features for each video, which allows us to generate a large number of high-quality proposals using BMN~\cite{lin2019bmn}.
Finally, we adopt the cascade scheme~\cite{cai2018cascade} to further fine-tune the proposals.

\subsection{Video Classification}
To improve the performance of video classification, we try to encode more temporal and spatial information.
The Slowfast~\cite{feichtenhofer2019slowfast} achieves excellent performance on video classification by decoupling spatial and temporal information in the temporal-spatial space.
Specifically, we choose Slowfast101 as our backbone, and each input clip has 32 frames by 15fps.
We first pre-train our backbone on Kinetics-600~\cite{carreira2018kinetics600} and then fine-tune it on the HACS dataset~\cite{zhao2019hacs}. 
The output of the network is a prediction of 200 categories. Because there is no background class, we only sample frames in segments that involve with action in training.
Finally, we add Batch Nuclear-norm~\cite{cui2020bnm} to the loss function to improve the generalization capabilities.

\begin{figure*}[t]
\centering
\includegraphics[width=16cm]{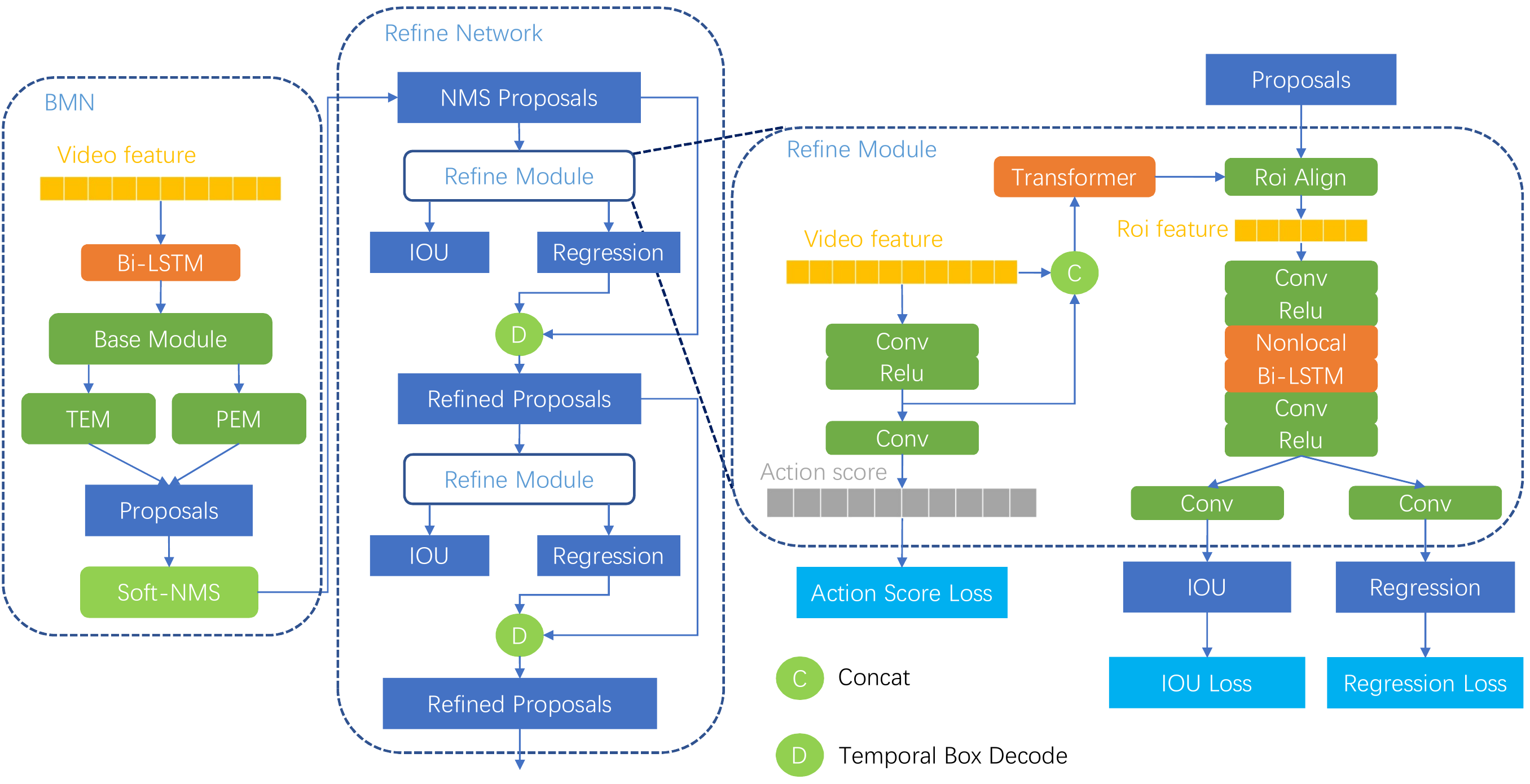}
\caption{Overview of our proposed framework. 
We first apply BMN to generate proposals and perform soft-nms on these proposals.
Then we input these proposals into the Refine Network to refine these temporal positions and re-predict IoU.
Actually, this is a standard cascade architecture. 
Then we embed a bidirectional LSTM layer in both BMN and the Refine Module to capture the forward and backward temporal information. 
In the Refine Module, we apply both Nonlocal and Transformer to obtain richer context information. 
At the same time, by adding the intermediate supervision of Action Score, the obtained features are more discriminative for local proposals.
}
\label{fig1}
\end{figure*}

\subsection{Generation of proposals}
\textbf{Features.} 
The fine-tuned model has a better representation of the HACS dataset. 
So we use the fine-tuned classification model's backbone to extract features for all videos. 
To encode more temporal information, we apply a dense sample strategy.
Specifically, we sample 32 frames with a 0.5s stride, and resize the short side of the clip to 256 pixels, and input the backbone to extract discriminative features. 
Then we take the features of stage ``Res5'' in Slowfast and use global average pooling to get a 2034-dimensional vector as the feature expression for each clip. 
Considering that every 32 frames represents a video length of about 2.13s, the overlaps between different clips will cause some frames to be repeatedly sampled.
However, same frames in different clips will have a different temporal context, so dense sampling can provide a more fine-grain feature representation, which can improve the quality of the generated proposals effectively.

\textbf{Boundary-Matching Network.}
In order to generate high-quality temporal proposals, Lin \emph{et.al}~\cite{lin2019bmn} proposed the Boundary-Matching Network(BMN).
BMN utilizes Temporal Evaluation Module (TEM) to generate temporal proposals.
Each proposal generated by TEM can map to BM confidence map which is generated by Proposal Evaluation Module (PEM), and provide confidence for proposals.
BMN  generates proposals with precise temporal boundaries as well as reliable confidence scores simultaneously.
For a video with a duration of $t$ seconds, we can use Slowfast101 to extract features with a temporal length of $T=round(2t)$.
In training and inference process of BMN, we resize the feature temporal length of each video to $D$. In our experiments we set $D$ to 200.
This is because the quality of the proposals is relatively high at a temporal length of 200 in the comparison of different temporal lengths.
LSTM can encode sequences very well, and in fact video also belongs to sequence information. 
We use a bidirectional LSTM to capture forward temporal information and backward temporal information of the video.
Our experimental results show that adding bi-directional LSTM to the base module of BMN can also improve performance.

\subsection{Fine-tuning of proposals}
\textbf{Refine Network.} 
BMN can generate a good deal of proposals. 
In the HACS dataset, most videos have long duration, while the temporal length of the above-mentioned features can be only set as a small number, \emph{e.g.}, 200 in our method, limited by the amount of calculation and memory.
Obviously, downsampling procedure will lose some temporal information. 
To solve this problem, we propose to further refine the proposals generated by BMN, which we call Refine Network.
We first perform soft-nms~\cite{bodla2017softnms} on the proposals, and sort them according to the corresponding scores. 
By this mean, a large number of redundant proposals can be removed, but only some good quality proposals would be fed for further retaining.
Then we input these proposals into the Refine Network. 
The Refine Module employs RoI Align Pooling~\cite{he2017mask-rcnn} technology to extract features for each proposal from the enhanced original video features.
By applying the features, we can further predict the corresponding IoU and temporal offsets of the proposals.
Then we repeat the Refine Modules several times to refine the proposals iteratively. In this way, a cascade~\cite{cai2018cascade} architecture is formed.
In our method, we embed three Refine Modules in the refine network.
As in ~\cite{cai2018cascade}, we set IoU thresholds for assigning positive labels to $0.5$, $0.6$ and $0.7$, respectively.

\begin{table}[t]
\footnotesize
\centering
\begin{tabular}{ccccc}
\toprule
Learning Rate & Remarks & Top1(\%) & Top5(\%)\\
\midrule
            0.001   &         &  89.71 & 98.11 \\
            0.001   &  +BNM    & 90.12 & 97.89 \\
            0.0005  &          & 91.19   & 98.98  \\
            0.0001  &          & 91.61   & 99.13  \\
            0.0001  &  +BNM    & 91.75   & 99.13 \\
            0.0001  &  +Transformer    & 91.86   & 99.06 \\
            0.0001  &  +BNM+Transformer    & 91.84   & 98.90 \\
            N/A    &  \makecell[c]{Ensemble \\ 14 models} & \textbf{94.32} & \textbf{99.68} \\
\bottomrule
\end{tabular}
\caption{Part of the results of the classification model. 
After ensemble, the model Top1 can obtain an absolute improvement to 2.36\%, which shows that there is a strong complementarity between different models. 
At the same time we found that BNM does not always improve performance. But it does not matter, what we need is the complementarity between different models.}
\label{tab:cls_result}
\end{table}

\begin{table}[t]
\footnotesize
\centering
\begin{tabular}{cccccc}
\toprule
Bi-LSTM & Nonlocal & \makecell{Action \\ Score} & Transformer &  AUC(\%)     &  mAP(\%)  \\
\midrule
        &         &              &              &   65.89  &   38.75 \\
\checkmark  &       &            &              &   65.75  &   38.90 \\
\checkmark  &  \checkmark     &            &    &   65.88  &   39.24 \\
\checkmark  &  \checkmark     &    \checkmark   &    &   65.88  &   39.48 \\
\checkmark  &  \checkmark     &    \checkmark   &  \checkmark  &   65.90  &   39.65 \\
\midrule
\multicolumn{4}{c}{ Ensemble} &                 \textbf{66.08} & \textbf{40.55} \\
\bottomrule
\end{tabular}
\caption{
Part of the results of Refine Network in the validation set. 
The addition of each module can improve the final mAP. 
The result of the ensemble of multiple models also shows that using different modules, the preferences of the models are different, and there is complementarity between the models. 
Our final ensemble result achieves \textbf{40.53} in terms of mAP(\%) on the test set.
}
\label{tab:map_result}
\end{table}

\begin{table*}
\centering
\begin{tabular}{cccccccc}
\toprule
Feature & $D$ &  Bi-LSTM & AR@1(\%) & AR@5(\%) & AR@10(\%) & AR@100(\%) & AUC(\%) \\
\midrule
I3D~\cite{carreira2017i3d}         & 200 &   0   & 18.26&37.94 & 47.47 & 70.89  & 60.99 \\
\midrule
\multirow{6}{*}{Slowfast~\cite{feichtenhofer2019slowfast}} & 256 &   0   & 19.91&41.04 & 50.80 & 73.47  & 64.10 \\
 & 200 &   0   & 19.87&41.15 & 50.92 & 73.10  & 63.83 \\
 & 180 &   0   & 19.89&41.16 & 50.99 & 73.22  & 63.91 \\
 & 160 &   0   & 19.80&40.98 & 50.65 & 72.60  & 63.49 \\
 & 200 &   1   & 20.31&41.26 & 50.72 & 72.31  & 63.44 \\
 & 200 &   2   & 20.25&41.33 & 51.08 & 72.61  & 63.73 \\
\midrule
\multicolumn{3}{c}{ Ensemble 10 models} & \textbf{20.83}	&\textbf{42.77}&\textbf{52.92}&\textbf{74.32}&\textbf{65.51}\\
\bottomrule
\end{tabular}
\caption{Part of the results of BMN.
We found that the results obtained using the Slowfast101 feature are significantly higher than obtained using the I3D feature. 
The test results of different temporal scales are not particularly different. But with the decrease of $D$, the performance is obviously reduced.
We noticed that the result of adding LSTM has decreased AUC. Nevertheless, AR@1 and AR@5 have significantly improved, which is very helpful for detection. 
In the ensemble process, we also found that the BMN result with LSTM is very complementary to other results, which can greatly improve the detection performance.}
\label{tab:bmn_results}
\end{table*}

\textbf{Refine Module.}
The refine module aims to further refine the candidate proposals and predict the corresponding IoUs.
Firstly, we predict score for each point on video feature whether there is an action, which we call Action Score, Similar to Actionness in ~\cite{lin2018bsn}.
Note that it is a supervised procedure to learn Action Score. 
By adopting the supervision information, the learned features can possess discriminative power.
%
Secondly, in order to predict IoU as accurately as possible, we should encode long-term temporal information.
%
It is obvious that local features can not know the actual duration of the action (especially when proposals intercept part of the action). 
%
Therefore, we need to extract features which have a global perception of videos.
In order to achieve the purpose, before performing RoI Align Pooling~\cite{he2017mask-rcnn}, we treat the video features as sequences and use Transformer~\cite{vaswani2017tansformer} in temporal dimension to obtain a global receptive field. 
Finally, we also use bidirectional LSTM network~\cite{hochreiter1997lstm} and Nonlocal~\cite{wang2018nonlocal} to enhance local features in the RoI features, which allows the network to fully integrate local temporal information.
%

\section{Experimental Results}
\subsection{Video Classification}
We choose Slowfast101 as the backbone of our classification model. 
In order to make the classification model more feature-rich, we train a variety of classification models based on this backbone. 
For example, different learning rates, adding Batch Nuclear-norm~\cite{cui2020bnm} in training, and adding Transformer~\cite{vaswani2017tansformer} to video features. 
While improving the results of a single model as much as possible, we also need the complementarity of features between different models. 
So even if the result of the single model is not the best, it can still improve the final classification results when performing ensemble procedure.
Finally, we fuse these classification results as our final classification results.
When there are $N$ classification models for ensemble, we set an adaptive weight for each model. 
These $N$ parameters are multiplied by the prediction results of each model before Softmax, and they will automatically converge to the value that makes the classification performance the best.
Some of our experimental results are shown in the Table~\ref{tab:cls_result}.

\subsection{BMN}
The purpose of BMN~\cite{lin2019bmn} is to initially obtain a large number of proposals for the fine network.
We also train many different models with different parameters or embedded modules to increase the complementarity between different results for ensemble scheme. 
For the ensemble of multiple BMN models, we first take the two $D$-dimensional vectors output by the TEM in each model, which are the action starting and ending probabilities, respectively. 
Then we take the two $D*D$ Boundary-Matching confidence maps output by PEM.
We perform weighted summation on the four maps between all models to obtain the results of model ensemble.
When multiple BMN models have different $D$, we resize the four maps to the same scale by linear interpolation.
The experimental results are shown in Table~\ref{tab:bmn_results}.

\subsection{Refine Network}
All Refine Networks in our experiments apply 3 Refine Module cascades.
During the inference, because each Refine Module will change the position of input proposals, so we only take the prediction result of the last layer of Refine Module as our final proposals.
We use the batch of proposals obtained after the BMN model ensemble as the input of $N$ different Refine Networks, and we can get $N$ batches of different adjusted proposals.
Finally, we use weighted summation to fuse these proposals as our final result. Our final detection result achieves $40.55\%$ on the validation set and $40.53\%$ on the test set in terms of mAP.
The experimental results are shown in Table~\ref{tab:map_result}.

\section{Conclusion}

In this technical report, we introduce the method designed for the HACS2020 competition.
The experiment results show that the proposals generated by BMN can also be further improved.
At the same time, for the fusion of temporal information, simply applying a larger convolution kernel to expand the receptive field does not effectively improve the quality of proposals. 
The effectiveness of LSTM, Nonlocal and Transformer shows that high-order information in temporal space is still an important research direction. 
In future works, we will further to explore for how to better encode temporal information.

{\small
\bibliographystyle{ieee}
\bibliography{egbib}
}

\end{document}